%% file: root.tex
\documentclass[letterpaper, 10pt, conference]{ieeeconf}

\IEEEoverridecommandlockouts
\usepackage{times}
\usepackage{amsmath,amssymb}
\usepackage{graphicx}
\usepackage{booktabs}
\usepackage{cite}
\usepackage{hyperref}
\usepackage{microtype}
\usepackage{tcolorbox}
\usepackage{xcolor}
\usepackage{colortbl}
\usepackage{orcidlink}
\definecolor{coachingblue}{RGB}{0, 102, 204}

\title{\LARGE \bf
Demonstration-Free Robotic Control via LLM Agents
}

\author{Brian Y. Tsui,
Alan Y. Fang,
Tiffany J. Hwu%
\thanks{All authors are independent researchers. Emails: brian.y.tsui@gmail.com, fangyenchi@gmail.com, tiffany.j.hwu@gmail.com}%
}

\begin{document}

\maketitle
\thispagestyle{empty}
\pagestyle{empty}

\input{sections/abstract}

\input{sections/background}

\input{sections/method}

\input{sections/evaluation}

\input{sections/limitations}

\input{sections/appendix}

\bibliographystyle{IEEEtran}
\bibliography{references}

\end{document}

%% file: sections/abstract.tex

\begin{abstract}

Robotic manipulation has increasingly adopted vision-language-action (VLA) models, which achieve strong performance but typically require task-specific demonstrations and fine-tuning, and often generalize poorly under domain shift. We investigate whether general-purpose large language model (LLM) agent frameworks, originally developed for software engineering, can serve as an alternative control paradigm for embodied manipulation. We introduce FAEA (Frontier Agent as Embodied Agent), which applies an LLM agent framework directly to embodied manipulation without modification. Using the same iterative reasoning that enables software agents to debug code, FAEA enables embodied agents to reason through manipulation strategies. We evaluate the unmodified Claude Agent SDK reasoning loop across the LIBERO, ManiSkill3, and MetaWorld benchmarks. With privileged environment state access, FAEA achieves success rates of 84.9\%, 85.7\%, and 96\%, respectively. This level of task success approaches that of VLA models trained with $\leq$100 demonstrations per task, without requiring demonstrations or fine-tuning, although observation modalities and evaluation protocols differ. With one round of human feedback as an optional optimization, performance increases to 88.2\% on LIBERO. This demonstration-free capability has immediate practical value: FAEA can autonomously explore novel scenarios in simulation and generate successful trajectories for training data augmentation in embodied learning.

\end{abstract}

%% file: sections/background.tex

\section{Motivation and Related Work}
\label{sec:background}

Learning-based robotic manipulation policies often require task-specific demonstrations, fine-tuning, and specialized execution pipelines. Despite the effectiveness of these policies, these requirements create substantial data-collection and engineering costs when adapting robots to each new task or environment. We therefore ask whether a general-purpose language model agent can instead perform manipulation using existing reasoning and tool-use capabilities, without demonstrations or robotics-specific training.

\subsection{LLM-Robot Integration via Custom Pipelines}
\label{sec:custom-pipelines}

A large body of prior work integrates language models into robotic manipulation through custom execution pipelines that translate language-level outputs into robot actions. These systems vary in their representations, planning abstractions, and feedback mechanisms, but typically rely on task- or platform-specific agent infrastructure.

\textbf{Planning and code generation approaches:} VoxPoser~\cite{huang2023voxposer} uses LLMs to generate composable 3D value maps for motion planning, achieving 70--90\% success on real-world tasks, but requires custom scene representations and environment-specific prompting. Code-as-Policies~\cite{liang2023code} generates executable robot code using hierarchical code generation to compose motion primitives, requiring handcrafted API abstractions for each platform. SayCan~\cite{ahn2022saycan} grounds LLM plans in robot affordances via learned value functions, achieving 84\% planning accuracy but necessitating pre-trained skill policies and custom scoring mechanisms.

\textbf{Feedback and multi-agent systems:} Inner Monologue~\cite{huang2022inner} introduced feedback loops where vision-language models provide scene descriptions to enable re-planning on failure. ProgPrompt~\cite{singh2023progprompt} extended this with programmatic prompting for situated task planning. More recently, MALMM~\cite{malmm2025} employs multi-agent LLM systems with separate Planner, Coder, and Supervisor agents for zero-shot manipulation.

\textbf{Embodied reasoning models:} Gemini Robotics~\cite{geminirobotics2025} exemplifies recent proprietary approaches to embodied AI, pairing a vision--language model with a smaller action model for robotic control and edge deployment. Their Gemini Robotics ER model extends Gemini 2.0 with enhanced spatial and temporal understanding, but remains closed-source and requires robotics-specific fine-tuning. FAEA instead tests whether an off-the-shelf agent's existing program-synthesis loop can discover task-level manipulation strategies without robotics-specific model training.

A common pattern across this literature is that each embodied agent system rebuilds agent infrastructure independently. Custom prompt engineering, error recovery mechanisms, system-specific tool interfaces, and execution loops are implemented separately for each platform. This fragmentation limits the transfer of engineering insights across systems. Meanwhile, many capabilities that embodied models are fine-tuned on, such as pointing or perspective understanding~\cite{geminirobotics2025}, are already present in open-source VLMs (e.g., Qwen3-VL~\cite{qwen3vl}, Molmo~\cite{molmo2024}) or can be addressed through traditional tools (camera-to-robot-frame transformations, inverse kinematics). Such capabilities can be exposed as agent tools.

In contrast to these modular approaches, end-to-end Vision-Language-Action (VLA) training represents the dominant paradigm in current robotics research. RT-1~\cite{brohan2022rt1} demonstrated that a Transformer-based architecture trained on 130,000 robot demonstrations could achieve 97\% success on over 700 training tasks. RT-2~\cite{zitkovich2023rt2} extended this by fine-tuning a 55-billion parameter vision-language model, achieving 62\% success on unseen scenarios. Following RT-2, numerous VLA variations have emerged along several directions: the $\pi$ series~\cite{black2024pi0, physicalintelligence2025pi05, physicalintelligence2025pi06} focuses on real-world teleoperation with intervention-based learning; NVIDIA's GR00T N1~\cite{nvidia2025groot} uses neural trajectory augmentation to expand training data $\sim$10$\times$ via synthetic video generation; and open-source efforts like OpenVLA~\cite{kim2024openvla} (7B parameters) and SmolVLA~\cite{smolvla2025} (450M parameters) have democratized access across model scales.

However, VLA training requires substantial data collection for each new task domain. Even with pre-training, new tasks and environments typically require fine-tuning with significant data, and models can remain brittle, with performance dropping from 95\% to below 30\% under modest environmental perturbations~\cite{fei2025liberoplus}. Contributing factors likely include the limited diversity of robot training data (4--5 orders of magnitude smaller than internet-scale text corpora~\cite{openx2023}), distribution shift between teleoperation-collected demonstrations and deployment conditions, and model sizes constrained by real-time inference requirements.

We ask a complementary question: can frontier Multimodal Large Language Model (MLLM) agents generate successful manipulation trajectories \emph{without} demonstration data, when given the ability to iterate through multiple trials? This reframes robotic manipulation by shifting emphasis from learning policies from demonstrations to discovering effective policies through iterative program synthesis. Relative to custom LLM-robot pipelines, FAEA retains a production agent SDK's native reasoning and error-recovery loop; relative to end-to-end VLAs, it discovers task programs through test-time iteration. We test three scoped claims: (1) one agent/model can discover manipulation programs without demonstrations; (2) the same prompt and agent loop transfer across simulators under privileged state and Cartesian control; and (3) successful programs can provide simulation trajectories. We do not claim generality across agent frameworks or physical hardware.

\subsection{From Software Agents to Embodied Agents}
\label{sec:agent-infrastructure}

Production agent frameworks share a common architecture rooted in the ReAct pattern~\cite{yao2022react}: a loop that interleaves reasoning with action. The agent receives a task, reasons about the next step, executes a tool call, observes the result, and adjusts its approach, repeating until completion. This core loop is conceptually simple, yet it provides exactly the control architecture needed for manipulation in partially observable embodied environments: observe, reason, act, observe result, adjust.

Around this simple core, production SDKs layer infrastructure for reliable long-running agents. Claude Agent SDK~\cite{claudeagentsdk} provides automatic context management for extended sessions, error handling with retry logic, and execution tracing for debugging. These capabilities are essential for iterative manipulation where agents may require dozens of attempts. Similar infrastructure exists in OpenAI Assistants API~\cite{openaiassistants}, Gemini ADK~\cite{geminiadk}, and LangChain~\cite{langchain}. Such production-grade capabilities are notably absent from robotics research prototypes. In this work, we use the Claude Agent SDK to provide agent infrastructure: the ReAct loop, context management across trials, and tool interfaces for robot control.

Moreover, recent benchmarking efforts have systematically evaluated frontier MLLMs for embodied control. EmbodiedBench~\cite{embodiedbench2025} evaluates 13 models across 1,128 tasks spanning household, navigation, and manipulation domains. Their findings reveal a critical distinction: native frontier MLLMs excel at high-level planning (Claude-3.5-Sonnet achieves 64--68\% on household tasks) but struggle with low-level manipulation (28.9\% for GPT-4o). FAEA therefore does not ask the MLLM to execute a real-time low-level control loop: the agent performs task decomposition and program refinement, while simulator-provided motion primitives and controllers execute Cartesian commands.

Despite this architectural alignment between software agents and embodied control, there remains limited systematic evaluation of whether production agent frameworks transfer to manipulation tasks. The robotics community has either built custom systems (assuming agent infrastructure must be robotics-specific) or focused on end-to-end VLAs (bypassing explicit agent reasoning entirely). Anthropic's Project Fetch~\cite{projectfetch2025} explored using Claude as a coding assistant for robot programming, with Claude-assisted teams completing tasks faster and writing significantly more code. However, they used Claude as a \emph{programming assistant} for humans, not as an autonomous agent directly controlling robots.

FAEA explores whether actively maintained frontier agent SDKs, with their structured tool interfaces, error handling, and execution tracing, are sufficient for embodied control and self-correction on practical tasks. We focus on task-level automation rather than real-time control: the agent reasons and acts at the timescale of seconds, delegating low-level motion to traditional controllers. Our evidence is limited to one SDK and model in simulation; other SDKs, sensing conditions, and physical systems remain unevaluated.

%% file: sections/method.tex

\section{FAEA Method Overview}
\label{sec:faea-overview}

The key insight of FAEA is that general-purpose agent frameworks designed for software engineering tasks can transfer to manipulation tasks in simulation without modifying the agent framework. We demonstrate this by using the Claude Agent SDK, the same infrastructure that enables Claude to write and debug code, to control robots in simulation environments. As discussed in Section~\ref{sec:agent-infrastructure}, we leverage the capability asymmetry between high-level reasoning and low-level control by delegating manipulation primitives to tools.

\paragraph{Terminology} We use the term \textit{demonstration-free} to describe FAEA, emphasizing that no expert demonstrations are required for task completion. Unlike single-attempt zero-shot evaluation, FAEA leverages native agent capabilities: the agent iteratively refines its approach based on execution feedback (error messages, success signals, visual observations) accumulated in context, without gradient updates. This is analogous to how humans learn manipulation through practice rather than observation. The agent discovers successful policies through trial and error at test time, averaging 2--26 attempts per task depending on difficulty (Table~\ref{tab:resources}).

\paragraph{Formal Framework}
Based on publicly available documentation of agent architectures~\cite{claudeagentsdk} and established patterns in coding agents, we conceptualize FAEA as an iterative program synthesis problem. Given task instruction $\ell$ and a tool set $\mathcal{T}$ (perception and control APIs), the agent generates a sequence of script attempts $\{\sigma_1, \sigma_2, \ldots, \sigma_N\}$ until success or termination. Each script $\sigma_i$ is a sequence of $T_i$ actions $(a_1, \ldots, a_{T_i})$ where each $a_t$ is an invocation of a tool from $\mathcal{T}$; execution produces a trajectory ending at terminal state $s_{T_i}^{(i)}$, evaluated by $\mathcal{E}: \mathcal{S} \rightarrow \{0, 1\}$.

Central to this iterative process, the agent maintains accumulated context $\mathcal{C}_i$ from previous attempts:
\begin{equation}
\mathcal{C}_i = \left\{(\sigma_j, o_j, \mathcal{E}(s_{T_j}^{(j)})) \mid j < i\right\}
\end{equation}
where $o_j$ denotes observations (error messages, visual feedback) from attempt $j$. Each new script is generated conditioned on instruction, tools, and history:
\begin{equation}
\sigma_i \sim \text{LLM}(\ell, \mathcal{T}, \mathcal{C}_i)
\end{equation}
The process terminates when: (1) success, $\mathcal{E}(s_{T_i}^{(i)}) = 1$, or (2) the agent concludes that further attempts are unlikely to succeed. No gradient updates occur, as the agent discovers successful policies through in-context program synthesis, explaining why manipulation success is achieved without robotic training data.

\paragraph{Prompt Template} Figure~\ref{fig:prompt-template} shows the FAEA prompt template that initializes the agent for each task. Following standard agent prompting practices, the template defines the agent's role, specifies success criteria, and structures expected outputs. The task instruction $\ell$ is populated into the \texttt{\{\{TASK\_DESCRIPTION\}\}} field. We evaluate two variants: \textit{baseline FAEA} uses the core template, while \textit{FAEA with coaching} augments it with high-level manipulation heuristics identified from analyzing failure cases in preliminary experiments. Figure~\ref{fig:faea-architecture} illustrates how the Claude Agent SDK orchestrates this ReAct loop.

\paragraph{Trace Validation} To ensure the agent solves tasks through legitimate reasoning rather than exploiting simulator vulnerabilities or brute-forcing solutions, we validate execution traces post-hoc. For each successful task, we run an automated review using Claude Code to flag suspicious patterns such as hardcoded coordinates copied from simulator internals, exhaustive grid searches, or solutions that bypass the intended task semantics. Manual inspection confirmed the two flagged MetaWorld traces, which were excluded (Section~\ref{sec:metaworld-eval}). We did not audit a random sample of unflagged traces using the same criteria, so the automated reviewer's false-negative rate remains unknown.

\begin{figure}[t]
\begin{tcolorbox}[
    colback=black!3!white,
    colframe=black!50!white,
    title=\textbf{FAEA Prompt Template},
    fonttitle=\bfseries\small,
    arc=2mm,
    boxrule=0.5pt,
    left=2pt,
    right=2pt,
    top=2pt,
    bottom=2pt,
]
\scriptsize\ttfamily
\textbf{\# Instruction}\\
You are an expert in robotics control. Your primary goal is to create a single script containing an episode that reaches success:\\
- with task: \{\{TASK\_DESCRIPTION\}\}\\[0.5em]

\textbf{\#\# Requirements}\\
- The final script must use success() as the metric\\
- Contains a sequence of actions\\
- Do not cheat or reference other tasks\\[0.5em]

\textbf{\#\# Guidelines}\\
- Follow the ReAct cycle\\
- Keep the final script simple\\[0.5em]

\textbf{\#\# Output}\\
- meta.json: success status and num\_tries\\
- episode.py: final script\\
- video.mp4: recording\\[0.5em]

{\color{coachingblue}\textbf{\#\# Coaching Tips}}\\
{\color{coachingblue}- Think about how a human would accomplish the task. Break it down into descriptive actions.}\\
{\color{coachingblue}- Build the episode iteratively in each ReAct cycle: record state, plan movements, review progress.}\\
{\color{coachingblue}- Keep a high z height when moving to avoid collisions.}\\
{\color{coachingblue}- Rotate gripper when it keeps hitting unintended objects.}\\
{\color{coachingblue}- Beware of obstacles in the action trajectory.}
\end{tcolorbox}
\caption{FAEA prompt template. Black text is used for baseline FAEA prompting; blue text shows coaching tips added for the enhanced variant. File paths and environment-specific details are masked.}
\label{fig:prompt-template}
\end{figure}

\begin{figure}[t]
    \centering
    \fbox{\parbox{0.95\columnwidth}{
    \small
    \textbf{FAEA Architecture with ReAct Loop}\\[0.5em]
    \begin{tabular}{@{}l@{}}
    \textbf{Input:} Task Instruction $\ell$ $\rightarrow$ Prompt Template $\rightarrow$ Claude Agent\\[0.3em]
    \hline\\[-0.8em]
    \textbf{ReAct Cycle} (repeat until success or budget exhausted):\\
    \quad 1. \textit{Reason}: Analyze task, plan approach\\
    \quad 2. \textit{Act}: Write/refine Python script $\sigma$\\
    \quad 3. \textit{Observe}: Execute in simulation, get $\mathcal{E}(s_T)$\\[0.3em]
    \hline\\[-0.8em]
    \textbf{Output:} Final script episode.py, success status
    \end{tabular}
    }}
    \caption{FAEA Architecture. The Claude Agent SDK orchestrates the ReAct loop: reasoning about the task, writing Python scripts, and observing execution results from LIBERO/ManiSkill simulations via Gymnasium interface.}
    \label{fig:faea-architecture}
\end{figure}

%% file: sections/evaluation.tex

\section{Simulation Evaluation}
\label{sec:simulation-eval}

We evaluate FAEA on LIBERO to establish native agent capability, then test whether human coaching provides additional optimization. We then apply the same agent loop and prompt to ManiSkill (domain randomization) and MetaWorld (different robot arm and environment). This tests transfer under a shared privileged-state and Cartesian-control abstraction, not other agent frameworks or physical hardware.

\subsection{Experimental Setup}
\label{sec:exp-setup}

\paragraph{Simulation Interface} Upon ingesting the FAEA prompt, Claude Agent autonomously discovers simulation APIs by reading documentation, following the same pattern it uses for software engineering tasks. Given access to Gymnasium documentation~\cite{towers2024gymnasium}, the agent identifies and uses standard interfaces of \texttt{step(action)}, \texttt{reset()}, \texttt{check\_success()}, and \texttt{get\_obs()} to write Python scripts that control the robot. The authors still configure each simulator and expose its API, so documentation use reduces task-program engineering but does not eliminate system-specific integration. LIBERO, MetaWorld, and ManiSkill all use \emph{absolute end-effector position control}, where actions specify target Cartesian coordinates; this common abstraction limits conclusions about transfer to other control interfaces.

\paragraph{Observation Modality and Comparison Context} FAEA accesses ground-truth state observations (object positions, gripper state) via \texttt{get\_obs()} rather than raw RGB images. This isolates whether frontier agents can discover successful manipulation strategies through iterative reasoning, separate from perception. VLA baselines in Tables~\ref{tab:libero}--\ref{tab:metaworld} use RGB observations. This comparison contextualizes FAEA's policy discovery capability. It does not claim equivalent performance under identical observation conditions.

We highlight two implications of this design choice. First, because simulation environments naturally provide state access, FAEA can serve as an automatic trajectory generator for VLA training data augmentation (see Section~\ref{sec:future}). Second, the decoupled evaluation aligns with recent work showing that control strategies learned with privileged state can transfer to real-world deployment when paired with appropriate perception pipelines~\cite{bestsimreal2025}.

\paragraph{Model Selection} We use Claude Opus 4.5 (\texttt{claude-opus-4-5-20251101}) for all experiments, despite being the most expensive model in the Claude family. Our rationale is twofold: (1) inference costs are becoming increasingly cost-effective for robotics applications, where NVIDIA's Vera Rubin platform promises 10$\times$ lower cost per token compared to current hardware~\cite{nvidiarubin2026}; and (2) we prioritize evaluating the best available model within our budget, using Anthropic's subscription access (\$200/month at time of writing). Evaluation throughput is constrained by subscription rate limits rather than cost.

\paragraph{Benchmarks} We evaluate on three complementary benchmarks:
\begin{itemize}
    \item \textbf{LIBERO}~\cite{liu2024libero}: 120 long-horizon manipulation tasks across four suites (Spatial, Object, Goal, Long-horizon) with Franka Panda robot.
    \item \textbf{ManiSkill3}~\cite{maniskill3_2024}: 14 manipulation tasks with domain randomization, testing generalization to visual and physical variations.
    \item \textbf{MetaWorld}~\cite{yu2020metaworld}: 50 tabletop manipulation tasks with Sawyer robot, testing generalization to a different robot arm and environment than LIBERO.
\end{itemize}
VLA baseline results for LIBERO and MetaWorld are taken from the SmolVLA publication~\cite{smolvla2025}; ManiSkill3 baselines are from ManiSkill3~\cite{maniskill3_2024}.

\subsection{Exploring FAEA Prompt Strategies on LIBERO}
\label{sec:libero-eval}

\begin{table*}[!htb]
\centering
\small
\begin{tabular}{@{}lcccccc@{}}
\toprule
\textbf{Method} & \textbf{Size} & \textbf{Spatial} & \textbf{Object} & \textbf{Goal} & \textbf{Long} & \textbf{Avg.} \\
\midrule
\multicolumn{7}{l}{\textit{FAEA (ours, demonstration-free, privileged state)}} \\
\quad Pilot (10 trials) & API & 70.0 & 100.0 & 60.0 & 68.9 & 70.6 \\
\quad Baseline (no limit) & API & 90.0 & 100.0 & 80.0 & 83.3 & 84.9 \\
\quad + coaching & API & 90.0 & 100.0 & 90.0 & 86.7 & 88.2 \\
\midrule
\multicolumn{7}{l}{\textit{VLA baselines (pre-trained, minimal LIBERO fine-tuning)}} \\
SmolVLA~\cite{smolvla2025} & 2.25B & 93 & 94 & 91 & 77 & 88.75 \\
SmolVLA~\cite{smolvla2025} & 0.45B & 90 & 96 & 92 & 71 & 87.3 \\
$\pi_0$~\cite{black2024pi0} (pretrained) & 3.3B & 90 & 86 & 95 & 73 & 86.0 \\
OpenVLA~\cite{kim2024openvla} & 7B & 84.7 & 88.4 & 79.2 & 53.7 & 76.5 \\
Diffusion Policy~\cite{chi2023diffusionpolicy} & -- & 78.3 & 92.5 & 68.3 & 50.5 & 72.4 \\
\midrule
\multicolumn{7}{l}{\textit{VLA with substantial LIBERO fine-tuning (30k+ demos)}} \\
$\pi_{0.5}$~\cite{physicalintelligence2025pi05} (fine-tuned) & 3B & 98.8 & 98.2 & 98.0 & 92.4$^\dagger$ & \textbf{96.85} \\
\bottomrule
\end{tabular}
\caption{LIBERO benchmark results (Franka Panda, 120 tasks). Task suites: Spatial (10), Object (10), Goal (10), Long (90). Demonstration-free FAEA baseline achieves 84.9\%; human coaching increases performance to 88.2\%. VLA baseline results are from SmolVLA. $^\dagger$The $\pi_{0.5}$ score uses the distinct 10-task LIBERO-10 suite, not FAEA's 90-task Long suite; its Long score and aggregate are therefore not task-matched comparisons to FAEA.}
\label{tab:libero}
\end{table*}

We systematically explore how different FAEA prompt strategies affect performance on LIBERO. Table~\ref{tab:libero} presents FAEA results alongside state-of-the-art VLA models.

\paragraph{Prompt Development} LIBERO contains 120 tasks. We use task 0 (LIBERO-Object) for prompt development: the agent first solves this task, and its successful script is included as an example in subsequent prompts. This avoids re-exploring LIBERO documentation for each new task, reducing token usage by $\sim$3$\times$ (measured on 10 LIBERO-Object tasks).

\paragraph{Pilot (10-trial cap)} We run a pilot study to establish feasibility with minimal prompting. A simple prompt describing the task goal is provided with the example script but without coaching, with a 10-trial cap per task. FAEA achieves \textbf{70.6\% success} across 120 tasks, demonstrating that demonstration-free manipulation is feasible with frontier agents. Performance is highest on object manipulation (100\%) and decreases for longer-horizon tasks (Goal: 60\%).

\paragraph{Baseline (no preset trial limit)} For the baseline condition, we remove the artificial trial cap and let the agent iterate until success or it concludes further attempts are unlikely to succeed. We retry the 35 tasks that failed in the pilot. The agent recovers \textbf{17/35 tasks (48.6\%)}, demonstrating that some pilot failures were due to the trial cap rather than fundamental incapability.

\paragraph{Human Coaching as Optimization} We test whether human-provided coaching can further improve baseline performance. We manually review video traces and add high-level tips to the prompt (e.g., maintaining gripper height, considering gripper orientation, avoiding obstacles). With coaching, recovery improves to \textbf{21/35 tasks (60.0\%)}, a \textbf{+11.4 percentage point} improvement over no-preset-limit trials alone.

FAEA baseline (84.9\%) is competitive with VLA models trained with limited LIBERO fine-tuning data, approaching $\pi_0$ pretrained (86.0\%) while requiring zero demonstrations. Human coaching increases performance to 88.2\%, approaching SmolVLA (88.75\%). These are contextual rather than controlled comparisons because observation modalities and inference protocols differ. The $\pi_{0.5}$ aggregate is not directly comparable because its fourth suite is LIBERO-10 rather than the 90 long-horizon tasks evaluated for FAEA.

\subsection{Demonstration-Free FAEA Outperforms Low-Data Tuning on ManiSkill3}
\label{sec:maniskill-eval}

\begin{table*}[!htb]
\centering
\small
\begin{tabular}{@{}llcccc@{}}
\toprule
\textbf{Method} & \textbf{Training} & \textbf{PickCube} & \textbf{PushCube} & \textbf{StackCube} & \textbf{PegInsertion} \\
\midrule
\multicolumn{6}{l}{\textit{FAEA (ours, demonstration-free)}} \\
\quad Baseline (no limit) & 0 demos & \textbf{100\%} & \textbf{100\%} & \textbf{100\%} & 0\% \\
\quad + coaching & 0 demos & \textbf{100\%} & \textbf{100\%} & \textbf{100\%} & 0\% \\
\midrule
\multicolumn{6}{l}{\textit{State observations (100 demos)}} \\
Diffusion Policy & 100 demos & 100\% & 95\% & 90\% & 38\% \\
ACT & 100 demos & 70\% & 99\% & 50\% & 14\% \\
BC~\cite{pomerleau1991bc} & 100 demos & 0\% & 69\% & 0\% & 0\% \\
\midrule
\multicolumn{6}{l}{\textit{RGB observations (100 demos)}} \\
Diffusion Policy & 100 demos & 76\% & 41\% & 61\% & 0\% \\
ACT & 100 demos & 28\% & 30\% & 33\% & 0\% \\
BC & 100 demos & 0\% & 0\% & 0\% & 0\% \\
\bottomrule
\end{tabular}
\caption{ManiSkill3 data efficiency comparison (4 representative tasks with available baselines; full 14-task results in Appendix). Baseline results are reported from ManiSkill3~\cite{maniskill3_2024}. On coarse manipulation tasks, FAEA matches or exceeds 100-demo trained models with zero demonstrations. Fine-grained precision tasks (PegInsertion) remain challenging for both approaches.}
\label{tab:maniskill-data-efficiency}
\end{table*}

This benchmark randomizes object positions, yaw rotations, and goal locations with each seed, testing whether demonstration-free performance holds under distribution shift. Table~\ref{tab:maniskill-data-efficiency} compares FAEA against imitation learning baselines trained on 100 demonstrations. Across 5 seeds and 14 tasks (70 total trials), FAEA achieves \textbf{85.7\%} success rate (60/70 tasks) in the baseline condition (no preset trial limit).

\paragraph{Per-Task Breakdown} Table~\ref{tab:maniskill-full} (Appendix) shows success rates across all 14 ManiSkill tasks.

\paragraph{Coaching as Negative Control} We apply the same LIBERO-derived coaching tips to ManiSkill as a negative control. Since ManiSkill tasks involve different objects, physics, and manipulation strategies, we expect these tips to be inapplicable. Indeed, coaching \emph{decreases} success rate from 85.7\% to \textbf{81.4\%} (57/70 tasks) while increasing API cost by 47\% (\$150 to \$220). This expected decrease confirms that coaching benefits are task-specific rather than universal, and that irrelevant coaching can actively harm performance.

FAEA achieves strong performance (85.7\%) on ManiSkill manipulation tasks with extensive domain randomization. Tasks requiring sub-centimeter precision (PegInsertion: 0\%, PlugCharger: 0--60\%) consistently fail; we analyze this limitation in Section~\ref{sec:limitations}.

\subsection{Cross-Embodiment Evaluation on MetaWorld}
\label{sec:metaworld-eval}

\begin{table}[!htb]
\centering
\small
\begin{tabular}{@{}lccccc@{}}
\toprule
\textbf{Method} & \textbf{Size} & \textbf{Easy} & \textbf{Med.} & \textbf{Hard} & \textbf{Avg.} \\
\midrule
\multicolumn{6}{l}{\textit{FAEA (ours, demonstration-free)}} \\
\quad Baseline$^\dagger$ & API & 100 & 90.9 & 96.9 & 96.0 \\
\quad + coaching & API & 100 & 100 & 100 & \textbf{100} \\
\midrule
\multicolumn{6}{l}{\textit{VLA baselines}} \\
SmolVLA & 2.25B & 87.1 & 51.8 & 67.0 & 68.2 \\
SmolVLA & 0.45B & 82.5 & 41.8 & 52.5 & 57.3 \\
$\pi_0$ & 3.3B & 80.4 & 40.9 & 40.4 & 50.5 \\
Diff. Policy & -- & 23.1 & 10.7 & 4.0 & 10.5 \\
\bottomrule
\end{tabular}
\caption{MetaWorld benchmark results (Sawyer, 50 tasks). FAEA achieves 96--100\% success, substantially outperforming all VLA baselines including SmolVLA (68.2\%). VLA baseline results are from the SmolVLA publication. $^\dagger$Two tasks marked failed despite succeeding (accessed internal state or used brute-force).}
\label{tab:metaworld}
\end{table}

MetaWorld uses a different robot arm (Sawyer vs.\ Franka Panda) and simulation environment, testing cross-embodiment transfer in simulation. With a simulator-specific interface configured, we apply the identical agent setup and prompts from LIBERO without task-specific adaptation. FAEA achieves \textbf{96\% success rate} (48/50 tasks) using native agent capabilities alone (Table~\ref{tab:metaworld}). We marked two tasks as failed even though the created tasks succeeded: one accessed internal MetaWorld state, and one performed an exhaustive search of final poses. FAEA achieves \textbf{100\% success} with LIBERO-derived coaching. These results concern similar simulated tabletop domains, not physical hardware.

\subsection{Computational Resources}
\label{sec:resources}

\begin{table*}[!htb]
\centering
\small
\begin{tabular}{@{}lcccccc@{}}
\toprule
\textbf{Category} & \textbf{Tasks} & \textbf{Cost/Task} & \textbf{Tokens/Task} & \textbf{Turns/Task} & \textbf{Tries/Task} & \textbf{Time/Task (min)} \\
\midrule
\multicolumn{7}{l}{\cellcolor{gray!15}\textbf{LIBERO} (Franka Panda, 120 tasks)} \\
\midrule
\quad Object & 10 & \$0.80 & 9.9K & 17.2 & 2.2 & 3.2 \\
\quad Long/90 & 90 & \$2.05 & 27.7K & 37.3 & 6.8 & 10.2 \\
\quad Spatial/10 & 10 & \$3.90 & 47.1K & 37.3 & 9.7 & 16.2 \\
\quad Goal & 10 & \$4.08 & 36.9K & 52.3 & 7.4 & 12.1 \\
\midrule
\multicolumn{7}{l}{\cellcolor{gray!15}\textbf{ManiSkill} (14 tasks $\times$ 5 seeds)} \\
\midrule
\quad Easy$^\dagger$ & 3 & \$0.51 & 6.5K & 15.4 & 1.1 & 2.0 \\
\quad Medium$^\ddagger$ & 7 & \$2.07 & 26.2K & 36.6 & 3.3 & 8.0 \\
\quad Hard$^\S$ & 4 & \$5.60 & 72.8K & 65.5 & 25.6 & 24.6 \\
\midrule
\multicolumn{7}{l}{\cellcolor{gray!15}\textbf{MetaWorld} (Sawyer, 50 tasks)} \\
\midrule
\quad Easy & 28 & \$0.94 & 11.2K & 23.4 & 1.4 & 3.7 \\
\quad Medium & 11 & \$1.26 & 15.7K & 28.1 & 1.0 & 3.4 \\
\quad Hard & 6 & \$1.19 & 15.3K & 27.4 & 1.3 & 5.7 \\
\quad Very Hard & 5 & \$1.35 & 16.4K & 27.8 & 1.0 & 2.5 \\
\bottomrule
\end{tabular}
\caption{Per-task computational resources by category. Cost/Task: API cost in USD. Tokens/Task: output tokens. Turns/Task: ReAct cycles. Tries/Task: simulation runs. Time/Task: wall-clock minutes. Difficulty strongly correlates with cost and time: ManiSkill Hard tasks cost 11$\times$ more than Easy ($\$5.60$ vs $\$0.51$) and take 12$\times$ longer (24.6 vs 2.0 min). $^\dagger$PickCube, PushCube, PullCube. $^\ddagger$LiftPegUpright, StackCube, PokeCube, RollBall, PlaceSphere, PickSingleYCB, TurnFaucet. $^\S$StackPyramid, PegInsertionSide, PlugCharger, AssemblingKits.}
\label{tab:resources}
\end{table*}

Table~\ref{tab:resources} reports mean end-to-end wall-clock times ranging from 2.0 to 24.6 minutes per task category, including repeated simulation attempts. These measurements represent iterative task discovery rather than policy inference and therefore are not directly comparable to VLA inference times. Reported API costs are estimates based on token usage and Claude Opus 4.5 pay-per-use pricing; the experiments used subscription-based access.

Computational cost scales with task difficulty across all benchmarks. On LIBERO, Goal tasks (60\% success) cost 5$\times$ more than Object tasks (100\% success): \$4.08/36.9K tokens vs \$0.80/9.9K tokens per task. Goal tasks also require 3$\times$ more turns (52.3 vs 17.2). On ManiSkill, Hard tasks cost 11$\times$ more than Easy (\$5.60/72.8K tokens vs \$0.51/6.5K tokens). This correlation reflects increased reasoning iterations for multi-step assembly and precision manipulation.

\subsection{Agent Reasoning and Problem-Solving Strategies}
\label{sec:agent-reasoning}

The quantitative results above establish \emph{what} FAEA achieves; here we examine \emph{how} the agent reaches successful manipulation policies. Analysis of execution traces reveals systematic reasoning patterns that mirror software debugging workflows.

\begin{table}[htb]
\centering
\small
\begin{tabular}{@{}lrrrrrr@{}}
\toprule
\textbf{Benchmark} & \textbf{Tasks} & \textbf{Bash} & \textbf{Write} & \textbf{Read} & \textbf{Web} & \textbf{Other} \\
\midrule
LIBERO & 180 & 48\% & 34\% & 14\% & 1\% & 3\% \\
ManiSkill & 139 & 55\% & 26\% & 14\% & 3\% & 2\% \\
MetaWorld & 100 & 56\% & 24\% & 14\% & 4\% & 2\% \\
\midrule
\textbf{Total} & 419 & 51\% & 30\% & 14\% & 2\% & 3\% \\
\bottomrule
\end{tabular}
\caption{Tool usage distribution across 18,473 total calls. Bash (script execution) dominates, followed by Write (script generation) and Read (output inspection). Web = WebFetch (documentation retrieval). Other = Grep, Glob, Edit, TodoWrite.}
\label{tab:tool-usage}
\end{table}

\paragraph{Tool Orchestration} FAEA leverages four primary tools: \texttt{Bash} (execute scripts), \texttt{Write} (generate code), \texttt{Read} (inspect outputs), and \texttt{WebFetch} (retrieve documentation). Table~\ref{tab:tool-usage} quantifies usage across 419 task executions totaling 18,473 tool calls. \texttt{Bash} dominates at 51\% as agents iteratively test hypotheses; \texttt{Write} (30\%) and \texttt{Read} (14\%) support the code-execute-inspect cycle. \texttt{WebFetch} accounts for only 2\%. Agents retrieve API documentation early (e.g., ManiSkill success criteria), then cache this knowledge in context for subsequent tasks. The consistent 14\% \texttt{Read} usage across benchmarks reflects the importance of output inspection in debugging.

\paragraph{Hypothesis-Driven Debugging} When initial attempts fail, the agent engages in structured root cause analysis rather than random retry. Consider LIBERO Task 10 (``put the black bowl on top of the cabinet''), which required 5 attempts. On the first failure, the agent observed that the bowl was not lifted:

\begin{quote}
\small\textit{``The bowl is moving when we approach it. Notice: Bowl starts at x=0.018, when EEF reaches target, bowl has moved to x=0.066. The gripper is pushing the bowl before it can grasp it.''}
\end{quote}

This observation led to a targeted hypothesis: the gripper's approach trajectory was displacing the object. The agent then wrote a debugging script to analyze finger tip positions relative to the bowl rim, discovering that the gripper needed to approach from directly above and lower straight down. This hypothesis-test-refine loop of examining coordinates, forming hypotheses, and testing with targeted scripts, is the same workflow a human robotics engineer would employ.

%% file: sections/limitations.tex

\section{Limitations}
\label{sec:limitations}

\paragraph{Precision Manipulation} Tasks requiring sub-centimeter precision (e.g., peg insertion, plug insertion) consistently fail across all conditions. Inspection of failure traces reveals that precision tasks fail due to insufficient positional accuracy in the ReAct loop: the agent correctly identifies the insertion goal but cannot achieve sub-millimeter alignment through discrete action commands. This suggests a fundamental mismatch between deliberative reasoning (seconds-scale) and precision control (milliseconds-scale). For such tasks, Vision-Language-Action models with real-time visual feedback may be more appropriate~\cite{geminirobotics2025}.

\paragraph{Latency} Full task discovery takes 2.0--24.6 minutes per task category on average (Table~\ref{tab:resources}), precluding real-time reactive control. FAEA is best suited for tasks tolerant of deliberative planning rather than reflex-like responses.

\paragraph{Evaluation Scope} This study evaluated a single agent framework (Claude Agent SDK) on a single model (Claude Opus 4.5); ablation studies across different agents and LLMs are required to establish framework-level generality. All experiments use privileged state and Cartesian control in simulation, so real-world and other control-interface validation remain future work.

\paragraph{Imperfect State Estimation} The debugging loop currently treats object poses and success signals as exact. On hardware, systematic pose bias could make successive scripts repeat the same incorrect correction, while stochastic noise could make the agent infer false progress or inconsistent failure causes. Errors can compound over multi-step tasks and are especially damaging near contact. A hardware implementation should attach uncertainty and timestamps to observations, reject stale or low-confidence estimates, obtain additional views when uncertain, and delegate final approach to closed-loop visual or force servoing. Evaluating success under injected pose noise, bias, latency, and missed detections is necessary before hardware deployment.

\section{Future Work}
\label{sec:future}

We highlight four forward-looking research directions:

\paragraph{Automatic Demonstration Generation for VLA Training} A key bottleneck in VLA training is the need for manual teleoperation to collect demonstrations, a labor-intensive process that scales poorly to novel tasks and environments. FAEA offers a potential solution: because the agent can autonomously discover successful manipulation policies in simulation without human input, it can generate demonstration trajectories across many novel scenarios automatically. This positions FAEA not as a replacement for VLA models, but as a complementary tool for scalable data generation in simulation. Downstream trajectory quality, diversity, and policy improvement remain to be evaluated.

\paragraph{Real-World Deployment} Deploying FAEA on physical robots will require perception modules to bridge the gap between simulation state access and real-world sensing. A perception layer must detect objects, estimate 6-DoF poses and uncertainty, and transform them through calibrated camera/robot frames. A motion layer must validate generated waypoints, solve collision-aware inverse kinematics, and execute them with closed-loop position or force control. An independent safety supervisor must enforce workspace, velocity, force, and retry limits. Each attempt would return synchronized observations and an independently computed success signal to the agent. These components require hardware-specific engineering and validation.

\paragraph{Characterizing Task Regimes for Agentic Control}
While this work focuses on manipulation tasks dominated by deliberative, task-level planning, an open question is how far agentic approaches like FAEA can extend toward more dynamic or reactive manipulation. Some tasks may admit hybrid strategies in which reflexive control is delegated to low-level controllers while higher-level sequencing and recovery are handled by the agent. Systematically characterizing which classes of tasks benefit from agentic reasoning, and where latency, reactivity, or continuous feedback become limiting factors, is an important direction for future work. Such analysis would clarify the boundary between tasks best served by agentic program synthesis and those requiring tightly coupled perception–action policies.

\paragraph{Guardrails and Safety} As shown in our MetaWorld experiments, powerful agent systems optimized for task success will explore any available path to achieve their goal when no hard boundaries exist, a capability that can produce dangerous behaviors in physical environments. Real-world deployment requires guardrails to constrain agent behavior, as persistent retry and creative problem-solving become hazardous around humans or fragile objects. FAEA execution traces are fully visible as Python scripts, enabling human review before deployment; however, readable traces do not by themselves guarantee safety or reviewer reliability.

\section{Conclusion}
\label{sec:conclusion}

We demonstrate that FAEA, a general-purpose frontier agent framework, achieves competitive manipulation performance without robotics-specific training. Across LIBERO, ManiSkill3, and MetaWorld, demonstration-free FAEA achieves high task success rates using native agent and MLLM capabilities under privileged state and a shared Cartesian action abstraction. The key insight is that iterative reasoning, the same capability that enables software agents to debug code, transfers to refining manipulation strategies through trial and error. These results establish feasibility for one SDK and model in simulation, not generality across frameworks or physical robots. Given its latency and sensing assumptions, FAEA is presently better supported as a simulation trajectory generator or deliberative layer above conventional control than as a standalone real-time robot controller.

\section*{Author Contributions}

B.Y.T. conceived the project and conducted experiments. A.Y.F. contributed to the experimental evaluation. T.J.H. contributed to manuscript editing and provided robotics expertise and guidance throughout the project.

\section*{Acknowledgment}

\textit{Special thanks to our tireless collaborator, Claude Code, which can automate the bulk of the research tasks.} Claude Code assisted with code generation, experimental iteration, and manuscript preparation. We found the tournament-style research prompting approach from AI Coscientist~\cite{boiko2023coscientist} particularly helpful for fact checking and refining experimental plans.

%% file: sections/appendix.tex

\appendix

\section{ManiSkill Per-Task Results}
\label{sec:appendix-maniskill}

\begin{table}[h]
\centering
\small
\begin{tabular}{@{}llcc@{}}
\toprule
\textbf{Task} & \textbf{Difficulty} & \textbf{No Limit} & \textbf{+ Coaching} \\
\midrule
PickCube-v1 & Easy & 100\% & 100\% \\
PushCube-v1 & Easy & 100\% & 100\% \\
PullCube-v1 & Easy & 100\% & 100\% \\
\midrule
LiftPegUpright-v1 & Medium & 100\% & 80\% \\
StackCube-v1 & Medium & 100\% & 100\% \\
PokeCube-v1 & Medium & 100\% & 100\% \\
RollBall-v1 & Medium & 100\% & 100\% \\
PlaceSphere-v1 & Medium & 100\% & 100\% \\
PickSingleYCB-v1 & Medium & 100\% & 100\% \\
TurnFaucet-v1 & Medium & 40\% & 60\% \\
\midrule
StackPyramid-v1 & Hard & 100\% & 100\% \\
PegInsertionSide-v1 & Hard & 0\% & 0\% \\
PlugCharger-v1 & Hard & 60\% & 0\% \\
AssemblingKits-v1 & Hard & 100\% & 100\% \\
\midrule
\textbf{Overall} & & \textbf{85.7\%} & \textbf{81.4\%} \\
\bottomrule
\end{tabular}
\caption{ManiSkill per-task success rates (5 seeds each). Coaching decreases overall performance, primarily due to PlugCharger where tips harm rather than help under domain randomization.}
\label{tab:maniskill-full}
\end{table}